\documentclass[final,5p,times,twocolumn,authoryear]{elsarticle}

\usepackage{amssymb}
\usepackage{amsmath}

\usepackage[utf8]{inputenc}
\usepackage[T1]{fontenc} 
\usepackage{natbib}
\usepackage{hyperref}
\usepackage{multirow}

\usepackage{xspace}
\newcommand{\cls}{\textsc{[cls]} token\xspace}

\usepackage{xcolor}
\usepackage{enumitem}

\makeatletter
\def\ps@pprintTitle{%
 \let\@oddhead\@empty
 \let\@evenhead\@empty
 \def\@oddfoot{}
 \let\@evenfoot\@oddfoot}
\makeatother

\begin{document}

\begin{frontmatter}

\title{Understanding Cell Fate Decisions with Temporal Attention}

\author[1,2,3,4]{Florian Bürger\corref{cor1}}
\ead{florian.buerger@uni-koeln.de} 

\author[1,2,3,4]{Martim Dias Gomes}
\author[5]{Adrián E. Granada}
\author[1,2,3,4]{Noémie Moreau}
\author[1,2,3,4]{Katarzyna Bozek}

\cortext[cor1]{Corresponding author}

\affiliation[1]{organization={Institute for Biomedical Informatics, Faculty of Medicine and University Hospital Cologne, University of Cologne},
            city={Cologne},
            country={Germany}}
            
\affiliation[2]{organization={Faculty of Mathematics and Natural Sciences, University of Cologne},
            city={Cologne},
            country={Germany}}

\affiliation[3]{organization={Cologne Excellence Cluster on Cellular Stress Responses in Aging-Associated Diseases (CECAD), University of Cologne},
            city={Cologne},
            country={Germany}}
            
\affiliation[4]{organization={Center for Molecular Medicine Cologne (CMMC), Faculty of Medicine and University Hospital Cologne, University of Cologne},
            city={Cologne},
            country={Germany}}

\affiliation[5]{organization={Granada Lab},
            city={Berlin},
            country={Germany}}

\begin{abstract}
Understanding non-genetic determinants of cell fate is critical for developing and improving cancer therapies, as genetically identical cells can exhibit divergent outcomes under the same treatment conditions. In this work, we present a deep learning approach for cell fate prediction from raw long-term live-cell recordings of cancer cell populations under chemotherapeutic treatment. Our Transformer model is trained to predict cell fate directly from raw image sequences, without relying on predefined morphological or molecular features. Beyond classification, we introduce a comprehensive explainability framework for interpreting the temporal and morphological cues guiding the model's predictions. We demonstrate that prediction of cell outcomes is possible based on the video only, our model achieves balanced accuracy of $0.94$ and an F1-score of $0.93$.
Attention and masking experiments further indicate that the signal predictive of the cell fate is not uniquely located in the final frames of a cell trajectory, as reliable predictions are possible up to $10$ h before the event. Our analysis reveals distinct temporal distribution of predictive information in the mitotic and apoptotic sequences, as well as the role of cell morphology and p53 signaling in determining cell outcomes. Together, these findings demonstrate that attention-based temporal models enable accurate cell fate prediction while providing biologically interpretable insights into non-genetic determinants of cellular decision-making. The code is available at \url{https://github.com/bozeklab/Cell-Fate-Prediction}.
\end{abstract}


\end{frontmatter}

\section{Introduction}
\label{sec:introduction}

A major challenge in cancer treatment is the partial response of tumors to chemotherapeutic drugs. Even when initially efficacious, such treatments often leave behind residual tumor cells, which can subsequently give rise to resistant clones and drive further disease progression. Understanding the mechanisms underlying cellular non-responsiveness to therapy is therefore of critical importance. Individual cells within identical population can exhibit markedly different outcomes under the same therapeutic conditions. While some cells undergo programmed cell death in response to treatment, others survive and continue proliferating, ultimately contributing to therapy resistance and relapse. Genetic factors play a key role in treatment response, however, a broad range of cell-intrinsic factors can lead to heterogeneous responses even among genetically identical cells \cite{albeck2008quantitative,spencer2009non,sharma2010chromatin}. These cells may differ in their epigenetic, transcriptomic, and proteomic states \cite{chakrabarti2018hidden}. Understanding the non-genetic factors that drive divergent outcomes is essential for improving treatment strategies and for gaining insight into fundamental cellular decision-making processes. Previous studies \cite{arora2017endogenous,chakrabarti2018hidden,korsnes2018single,wolff2018inheritance} have identified multiple molecular factors influencing cell fate following treatment. In addition to markers such as p53, these studies highlight the importance of dynamic cellular properties, including cell-cycle phase and proliferative state, in determining drug susceptibility. Although these studies inspected the link of individual factors to cell fate, a systematic investigation of the predictability of cell fate under chemotherapeutic treatment given various dynamical factors is still lacking.

Deep learning methods provide a powerful means to investigate not only molecular determinants of cell fate, but also morphological and spatial aspects of the cellular environment. Rather than relying on the manual selection and quantification of predefined features \cite{el2014temporal}, deep learning approaches enable direct assessment of cell fate predictability using raw time-lapse microscopy image data. This allows simultaneous capture of dynamic, morphological, and contextual features of cells within a culture. 

Recent studies have adopted representation learning techniques to capture cellular dynamics directly from raw data. Several self-supervised approaches leverage autoencoder-based architectures to learn latent representations of cell morphology and dynamics, which are subsequently analyzed in search for their biological relevance. For example, \cite{wu2022dynamorph} learned representations from time-lapse sequences and related principal components of the latent space to handcrafted morphological features, enabling post-hoc interpretation. Similarly, \cite{ulicna2023learning} employed vector-quantized autoencoders to learn discrete representations for cell-cycle classification. While these methods provide useful representations of cellular dynamics, they primarily focus on cell-cycle stage rather than direct prediction of divergent outcomes of cells under treatment.

\cite{yamamoto2022probing} explored cell fate prediction in multicellular contexts. The study proposed an interpretable graph neural network framework to model cell fate decisions in tissues, explicitly incorporating cell--cell interactions through graph structures derived from live imaging data. Their work highlights the importance of spatial context and collective dynamics but does not consider intracellular signaling dynamics in the prediction.

\cite{cunha2025ai4cellfate,soelistyo2022learning} introduced autoencoder-based cell representations to predict cell fate. Latent representations were analyzed for their link to the morphological and intensity-based features and predictability. 

Among these studies, a systematic investigation of when predictive information emerges over time and how temporally localized cues relate to interpretable cellular features remains limited. In particular, none of the studies explored the predictive power of Transformer models and their interpretability via the attention mechanism. Transformers represent powerful architectures to process sequential data, such as time-lapse recordings. Their use in the study of cellular dynamics, particularly in the context of chemotherapeutic treatment and dynamic molecular reporters, has not been demonstrated yet. 

Here, we develop a deep learning model to predict the fate of individual cancer cells following exposure to chemotherapeutic treatment. Specifically, we propose a Transformer-based architecture that operates on temporal sequences of center-cropped image patches extracted from long-term time-lapse microscopy data. The input captures not only cellular morphology but also the dynamic activity of multiple transiently active molecular markers, including p53, circadian, and cell-cycle reporters. Importantly, the model predicts whether a cell undergoes apoptosis or mitosis directly from raw image sequences, without relying on predefined morphological features, allowing it to learn task-relevant representations in a data-driven manner.

We investigate not only the extent to which cell fate can be predicted from video-encoded cellular trajectories but also develop model explainability strategies to address two key questions: (1) when during a cell’s lifetime fate decisions become predictable, and (2) which signals are most informative for those decisions.
Our work demonstrates that the combination of rich time-lapse microscopy data, temporal modeling, and attention-based analysis enables the extraction of meaningful biological signals and facilitates the discovery of determinants of cell fate following chemotherapeutic treatment.

The contributions of this work can be summarized as follows:
\begin{itemize}
    \item We introduce a Transformer-based model for predicting cell fate from long-term single-cell raw image sequences.
    \item We demonstrate prediction accuracy on a challenging, imbalanced live-cell imaging dataset and show that prediction of cell fate is indeed possible from videos of cell trajectories.
    \item We perform temporal truncation experiments to assess how predictive information is distributed across time.
    \item We propose an attention-based explainability framework that links temporally important frames to interpretable morphological and intensity-based features.
\end{itemize}
\section{Methods}
\label{sec:methods}

\begin{figure*}
    \centering
    \includegraphics[width=\linewidth]{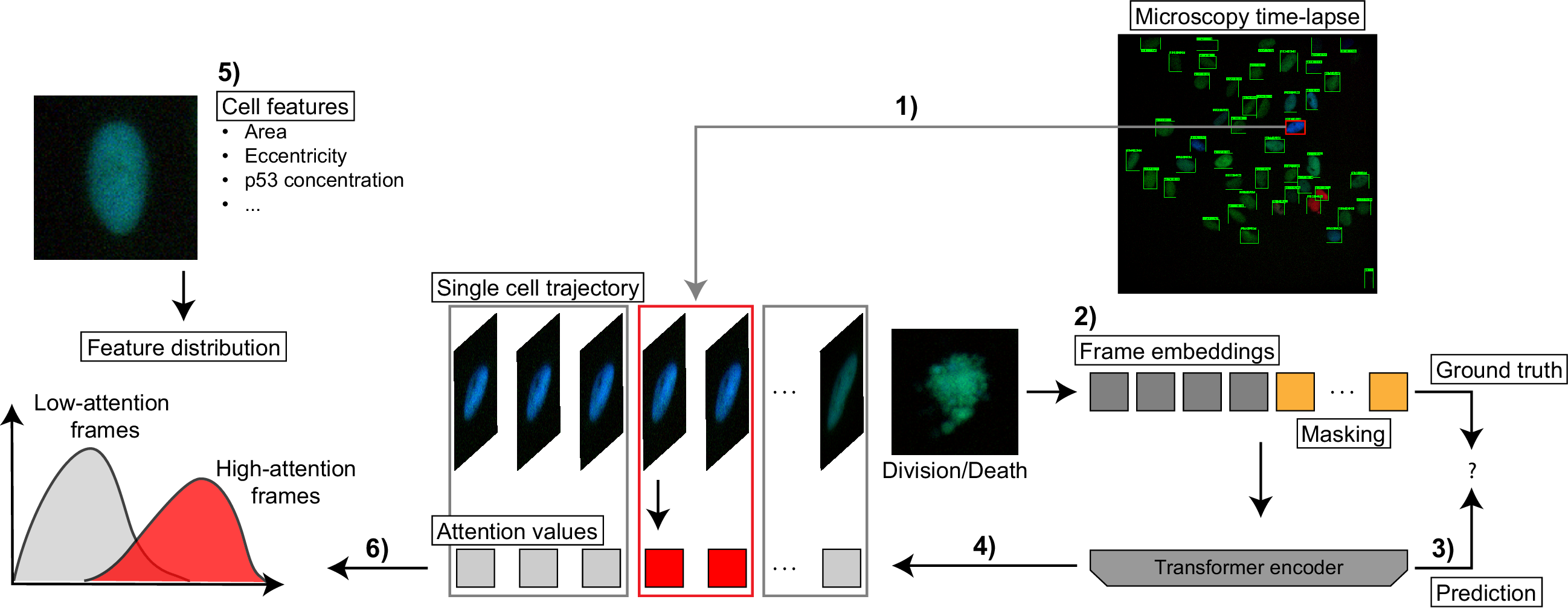}
    \caption{Overview of the proposed cell fate prediction framework and the explainability using temporal attention analysis. 
    \textbf{1)} From time-lapse microscopy recordings, we extract single-cell trajectories. Each trajectory starts at the first appearance of a cell and ends at the last frame prior to the fate event (division or death). 
    \textbf{2)} For each cell patch, frame-level embeddings are computed using a ResNet-50 backbone. We randomly mask a portion of frames at the end of each trajectory during training. 
    \textbf{3)} The sequence of frame embeddings is processed by a Transformer encoder, which integrates temporal information to predict the final cell fate. 
    \textbf{4)} To enable interpretability, we extract the attention weights assigned to individual frames and categorize them into high-attention and low-attention frames. 
    \textbf{5)} In parallel, we compute handcrafted cell features for each patch (e.g., area, eccentricity, p53 concentration). 
    \textbf{6)} Finally, we compare the feature distributions of high- and low-attention frames to identify biologically meaningful patterns associated with the model’s decision-making process.}
    \label{fig:mainfig}
\end{figure*}

\subsection{Dataset}
\label{subsec:methodology:dataset}

We use a publicly available live-cell imaging dataset \cite{burger2025transienttrack}, comprising $274$ time-lapse fluorescence microscopy videos. Each video consists of $186$ frames with a spatial size of $1024 \times 1024$ pixels, acquired every $30$ minutes over a continuous period of $93$ hours.

Out of the $274$ videos, $118$ capture cell cultures under treatment with the chemotherapeutic agent cisplatin. Cisplatin induces DNA damage and was administered $48$ hours after the start of imaging at three different dosage levels (high, medium, and low). Three fluorescent reporters were used to monitor key cellular pathways: CFP-hGeminin, which reports on cell-cycle progression by indicating S/G2/M phases; NR1D1::VNP, a fluorescent circadian clock and metabolism reporter reflecting cellular stress and circadian rhythm activity; and a p53 transcriptional activity reporter, capturing stress- and DNA-damage–related signaling dynamics. Together, these channels provide a rich temporal record of molecular signaling and cellular state transitions. 

In each video, cells are tracked and assigned unique cell identifiers. Events of mitosis and apoptosis are labeled. In total, the dataset contains approximately $51{,}000$ individual cell trajectories (averaging $\sim$187 cells per video), including about $18{,}000$ mitotic events and $3{,}500$ apoptotic events.

To predict fate of individual cells, we extracted temporal image sequences by center-cropping each annotated cell at a spatial resolution of $112 \times 112$ pixels around its centroid, ensuring that the entire cell body was captured throughout the sequence. Each sequence was paired with its corresponding cell fate label - mitotic vs. apoptotic outcome. For our approach, we exclusively included cells with annotated mitotic or apoptotic fates. Furthermore, we included only sequences longer than $10$ frames. When a crop extended beyond the image boundaries, zero-padding was applied to preserve a fixed patch size. The raw dataset is published at \url{https://doi.org/10.4126/FRL01-006526608}. The script used to generate the cropped temporal sequences from the raw dataset is provided in our aforementioned GitHub repository.

\subsection{Classification}
\label{subsec:methodology:architecture}

An overview of the proposed Transformer-based cell fate prediction framework and the attention-guided analysis pipeline is shown in Fig.~\ref{fig:mainfig}.

Each input consists of a time-lapse sequence of cropped single-cell images. Individual cell images are first processed independently by a ResNet50 backbone to extract frame-wise (\(2024\)-dimensional) feature embeddings. These embeddings represent input tokens to our Transformer model. \cls is prepended to the sequence of image tokens to serve as a global representation of the entire trajectory.

To encode temporal order, we add sinusoidal positional embeddings \cite{vaswani2017attention} to all elements of the sequence. The resulting representation is then input into a Transformer encoder composed of $8$ identical layers, each employing multi-head self-attention with $4$ attention heads. This encoder models long-range temporal dependencies and interactions between time points across the full cell trajectory.

On top of the encoder output, we apply an additional attention layer to aggregate temporal information and capture the relative relevance of individual frames for the final prediction. Layer normalization is applied both before and after this attention layer to stabilize optimization and improve representational consistency.

Finally, the transformed \cls is passed through a linear classification head to predict the cell fate based on the full temporal sequence.

\subsection{Explainability}
\label{subsec:methodology:explainability}

In addition to predicting cell fate, we investigate which frames within a temporal sequence are the most influential for the model’s decision-making process and how these frames differ from the less informative ones. To this end, we leverage the attention mechanism of the model as a proxy for temporal importance.

During inference, we extract attention weights from the final attention layer of the network. Given a sequence $s$ of length $N$, we denote the attention weights assigned to individual frames $\{f_{s,i}\}_{i=1}^{N}$ by $\{a_{s,i}\}_{i=1}^{N}$. To identify the most relevant frames within a sequence, we define a sequence-specific threshold $\tau_s$ as the $90$th percentile of the attention weights of $s$. Using this threshold, we partition the set of frames into two disjoint subsets:
\begin{equation*}
    \mathcal{H}_s := \{f_{s,i} \mid a_{s,i} \geq \tau_s\}
\end{equation*}

\begin{equation*}
    \mathcal{L}_s := \{f_{s,i} \mid a_{s,i} < \tau_s\}
\end{equation*}
Thus, $\mathcal{H}_s$ contains the $10\%$ of frames within sequence $s$ receiving the highest attention weights, while $\mathcal{L}_s$ comprises the remaining frames.

To search for distinguishing features between high- and low-attention frames, we quantify morphological and intensity-based characteristics of cells in all video frames. Individual cells are segmented using Cellpose \cite{stringer2021cellpose}, and visual features are computed from the resulting segmentation masks using \textit{scikit-image} \cite{van2014scikit}. To avoid confounding effects from neighboring cells appearing in the cropped patches, only the centrally located cell is considered for feature computation.

The extracted features include cell area, perimeter, equivalent diameter, eccentricity, solidity, circularity, and the mean intensity of each fluorescence channel. 

For each sequence and feature, we conduct permutation tests with $50{,}000$ iterations to assess whether feature distributions differed significantly between $\mathcal{H}_s$ and $\mathcal{L}_s$. This yields one p-value per feature and sequence. In addition to statistical significance, we quantify the magnitude and direction of differences using Cliff’s delta \cite{cliff1993dominance}. Unlike Cohen’s $d$, Cliff’s delta does not assume normality or equal variances and is robust to skewed distributions and unequal sample sizes \cite{romanoAppropriateStatisticsOrdinal2006}.

Across all sequences, this results in $n$ p-values and $n$ corresponding Cliff’s delta values per feature. To assess global consistency of within-sequence effects, we additionally compute the median Cliff’s delta and its $95\%$ confidence interval. A median effect size whose confidence interval is shifted away from zero is interpreted as evidence of a consistent effect, with the magnitude indicating the strength and direction of the association.

\subsection{Training}
\label{subsec:methodology:training}

We randomly split our cell trajectories into training, validation, and test sets using an $0.8$/$0.1$/$0.1$ ratio, while approximately preserving the distribution of mitotic and apoptotic events across all splits. During training, we used a batch size of $4$ and to address the pronounced class imbalance, we employed a weighted random sampler to approximately balance mitotic and apoptotic samples. As cell trajectories vary in temporal length, sequences were padded at the end of the sequence with empty tokens to match the maximum sequence length within each batch. During training, padded positions were masked within the Transformer encoder to ensure that they do not contribute to the attention computation.

To improve generalization, we applied a range of data augmentations independently and stochastically during training. The augmentation pipeline includes random geometric transformations, such as horizontal and vertical flips, as well as appearance-based augmentations including color jittering and Gaussian blurring. In addition, we applied a sequence-level augmentation by randomly masking a contiguous segment at the end of each sequence, with the masked portion ranging between $10\%$ and $50\%$ of the total sequence length. Each augmentation is applied with a fixed probability, increasing the diversity of the training data and improving model robustness.

The ResNet backbone, Transformer encoder, final attention layer, and classification head were trained from scratch using binary cross-entropy loss. Optimization was performed using the Adam optimizer with a learning rate of $1 \times 10^{-5}$ and a weight decay of $1 \times 10^{-3}$. Training was conducted for up to $86$ epochs with early stopping based on the validation loss, and the model parameters corresponding to the best validation performance were retained. All experiments were conducted on a system equipped with an AMD EPYC 7662 CPU, $150$~GB of RAM, and an NVIDIA Tesla A100-SXM4 GPU with $40$~GB of memory. Reproducibility was ensured through fully deterministic training and evaluation. All experiments were run with fixed random seeds across Python, NumPy, and PyTorch, and deterministic computation settings were enforced to guarantee identical results across runs.

\section{Experiments and Results}
\label{sec:experiments}

In this section, we present experimental results evaluating both the predictive performance and explainability of our model for single-cell fate prediction. We first quantify classification accuracy across different temporal ranges. We then leverage attention-based explainability analyses to gain insights into the temporal signals and cell characteristics that drive the model's predictions.

\subsection{Classification}
\label{subsec:experiments:classification}

Classification performance was evaluated on a held-out test set of cell trajectories. The test set comprised $2{,}089$ sequences, including $439$ apoptotic and $1{,}650$ mitotic events, reflecting a strong class imbalance of the entire dataset.

\begin{figure}
    \centering
    \includegraphics[width=0.7\columnwidth]{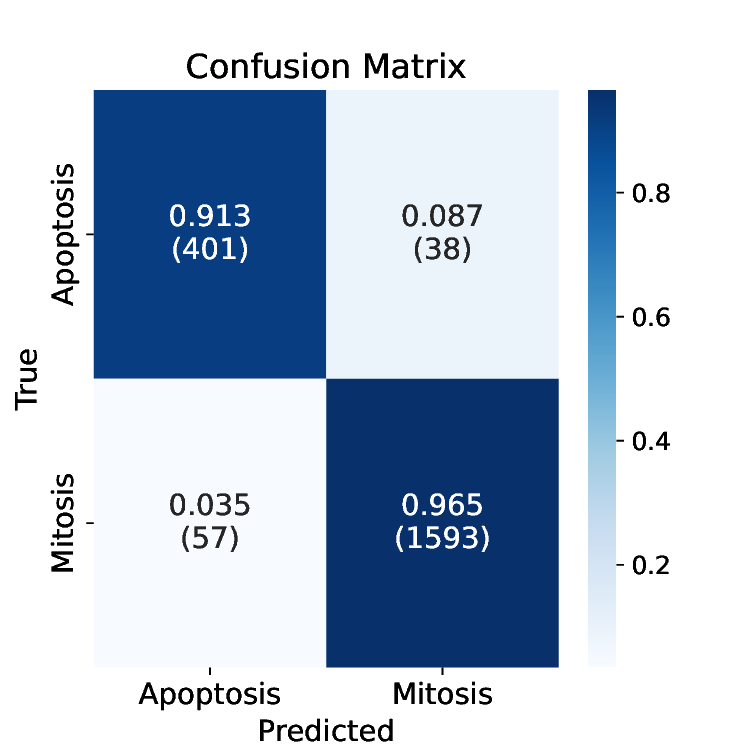}
    \caption{Confusion matrix normalized over the ground-truth labels, with absolute counts shown in parentheses.}
    \label{fig:results:confusion}
\end{figure}

Our model achieved a balanced accuracy (BAcc) of $0.94$ and an F1-score of $0.9326$ on the test set. As shown in the confusion matrix in Fig.~\ref{fig:results:confusion}, the proposed approach correctly identified $91.3\%$ of apoptotic samples ($401$ out of $439$), demonstrating a high recall for the minority class. Only $8.7\%$ of apoptotic instances were misclassified as mitotic events.

At the same time, the model maintained a low false-positive rate for apoptosis prediction, with only $3.5\%$ of mitotic samples incorrectly predicted as apoptotic. This indicates that the high recall for the minority class was not achieved at the expense of excessive number of false positives. Overall, these results demonstrate that the proposed model reliably discriminates apoptotic from mitotic events despite the pronounced class imbalance in the dataset.

\subsubsection{Temporal Truncation}
\label{subsubsec:experiments:classification:truncation}
For all further experiments and analyses, we restricted the evaluation to sequences originating from videos in which cisplatin was administered. This enables us to specifically study temporal cues that are predictive of divergent cell fate decisions under chemotherapeutic treatment, while avoiding confounding effects from untreated control conditions. 

\begin{figure*}[t]
    \centering
    \includegraphics[width=\textwidth]{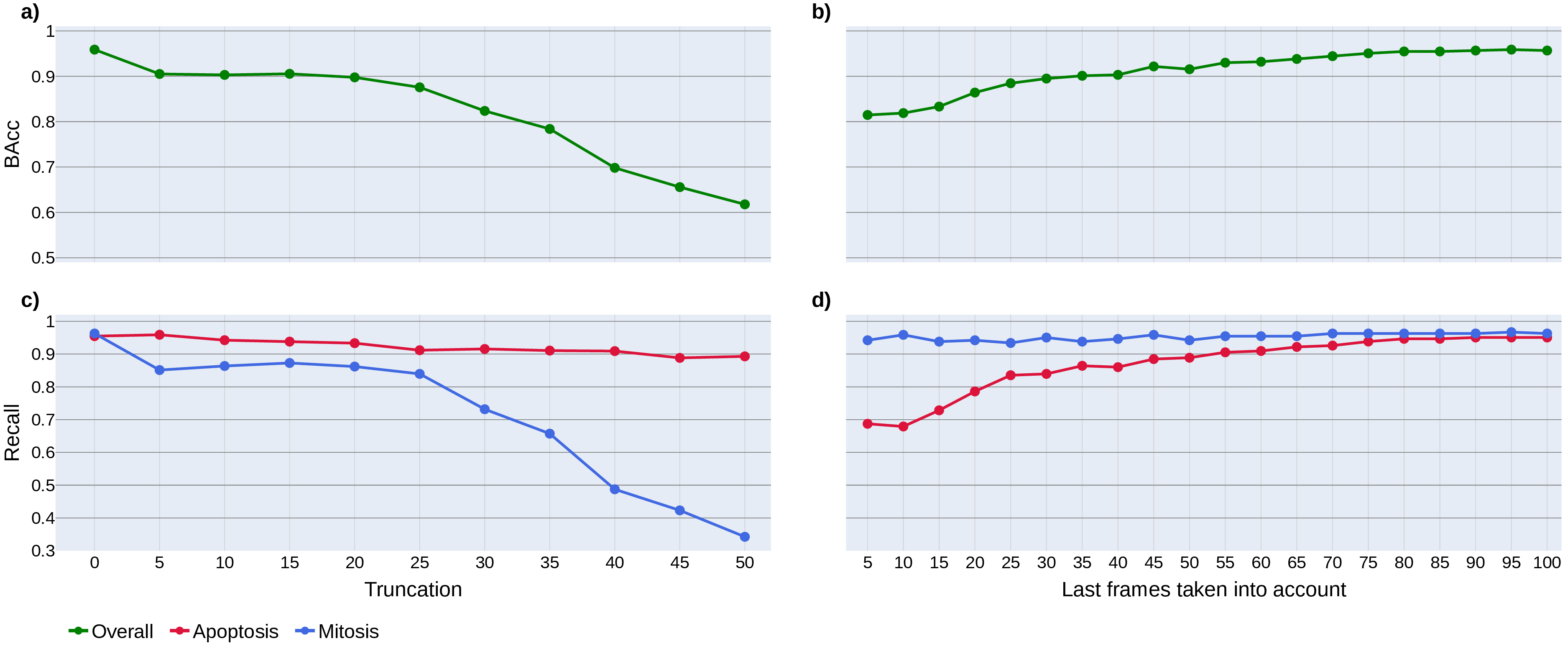}
    \caption{Temporal truncation analysis to assess the importance of late parts of the cell trajectory for its fate prediction. 
    \textbf{a)} Balanced accuracy when progressively truncating sequences by removing an increasing number of frames from the end of each trajectory. 
    \textbf{b)} Balanced accuracy when restricting the model input to only the last $k$ frames of each sequence. 
    \textbf{c)} Class-wise recall under progressive truncation from the sequence end, reported separately for apoptosis and mitosis. 
    \textbf{d)} Class-wise recall when predictions are based solely on the last $k$ frames, highlighting the contribution of late temporal information for each cell fate.}
    \label{fig:results:truncation}
\end{figure*}

Since a cell might undergo visual changes shortly before its death or division that clearly reveal its fate, we next performed an experiment in which we removed an increasing number of frames at the end of each sequence and inspected the accuracy of prediction on this way shortened cell trajectories. We evaluated the model on the corresponding test set while progressively truncating the final $0, 5, 10, \ldots, 50$ frames of each sequence. Sequences shorter than the respective truncation length were excluded from this analysis. The resulting performance is shown in Fig.~\ref{fig:results:truncation}a.

When evaluating our model on the cisplatin-treated test set, we observed a BAcc of $0.958$. Omitting the final $5$ frames resulted in a moderate performance decrease to a BAcc of $0.905$. Performance remained relatively stable when up to $20$ frames were omitted (BAcc: $0.898$). 
In contrast, truncating $25$ or more frames led to a pronounced decline in performance, which continued steadily with increasing truncation length, reaching a BAcc of $0.618$ when $50$ frames were removed. 
Class-specific recall values (Fig.~\ref{fig:results:truncation}c) further reveal distinct truncation sensitivities for the two cell fate outcomes. Recall of apoptotic sequences remained comparatively stable across truncation levels, decreasing only by $0.062$ from $0.955$ without truncation to $0.893$ when the final $50$ frames were removed. In contrast, recall of mitotic sequences showed a sharp drop when truncating just $5$ frames, decreasing from $0.963$ to $0.851$. Thereafter, mitotic recall remained relatively stable until truncating approximately $25$ frames (recall: $0.839$), followed by a substantial decline to $0.342$ when the final $50$ frames were removed. 
These results indicate that the temporal distribution of predictive signal differs between the two cell fate classes. In particular, mitotic sequences appear to be predictable from signals late in the sequence, with the final $\sim 13$ hours prior to the event contributing most strongly to discriminative performance, and earlier time points providing only limited predictive information. In contrast, apoptotic sequences which are predictable from signals across a broader temporal window.

In a complementary analysis, we examined how many frames at the end of a sequence were sufficient to predict the cell fate. Instead of truncating the end of the sequence, we retained only the last $k$ frames and discarded all preceding frames. Fig.~\ref{fig:results:truncation}b shows the corresponding results. Using only the last $k=5$ frames yielded a BAcc of $0.815$, which increased steadily as more frames were included. Near-optimal performance was achieved when using approximately the last $k=80$ frames (BAcc: $0.955$), after which additional frames provided only marginal improvements. 

Here, we also analyzed class-specific performance by computing recall for each event type while restricting the model to only the last $k$ frames. Fig.~\ref{fig:results:truncation}d shows that mitotic sequences could be reliably identified using only the last $k=5$ frames, achieving a recall of $0.942$. Recall increased slightly when considering the last $k=10$ frames ($0.959$), followed by a decrease at $k=15$ ($0.938$). Subsequently, recall stagnated until $k=65$. Only using the last $k=70$ and more resulted in a recall greater than $0.96$. This shows, that only taking the last $5$ to $10$ frames suffices for the model to predict mitotic events. 
In contrast, apoptotic sequences required substantially more temporal context for reliable prediction. When restricting the sequences to the last $k=5$ frames, recall was at $0.687$. It increased progressively as additional frames were included, reaching $0.946$ at $k=80$ frames, after which further increases in sequence length did not yield additional performance gains. 
Again, this suggests that signals predictive of cell division appear shortly prior to the event while signals predictive of cell death can be distributed over longer time spans.

\subsection{Explainability}
\label{subsec:experiments:explainability}

Beyond predictive performance, understanding how and when the model arrives at its decisions is critical for model interpretation. In the following, we analyze the model's internal attention mechanism to investigate (1) when during a cell's trajectory predictive information is most strongly utilized and (2) which features contribute most to fate discrimination.

\subsubsection{Attention Focus}
\label{subsubsec:experiments:explainability:attention_focus}

To corroborate the findings based on sequence truncation and to identify what are the signals predictive of particular cell fate, we next performed attention-based explainability analysis. Following the procedure described in Sec.~\ref{subsec:methodology:explainability}, we identified the high-attention and low-attention frames $\mathcal{H}_s$ and $\mathcal{L}_s$ in all sequences $s \in S$.

\begin{figure*}[t]
    \centering
    \includegraphics[width=\textwidth]{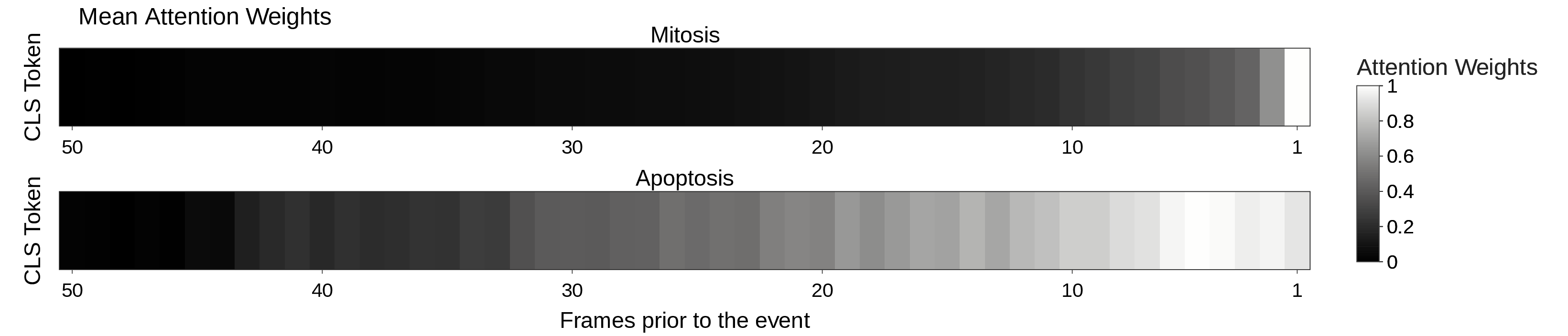}
    \caption{Aggregated attention weights across correctly classified sequences, aligned to the final frame and normalized by the global minimum and maximum attention values.}
    \label{fig:analysis:attention}
\end{figure*}

In Fig.~\ref{fig:analysis:attention}, we aggregated attention weights across all sequences by aligning them to the final frame of each sequence. Specifically, we considered up to $50$ frames preceding the cell event and computed the mean attention weight per frame. Attention values were normalized using the global minimum and maximum attention weights. Only sequences that have been correctly classified by the model were included in this analysis.

In sequences of dividing cells, the aggregated attention profile shows a pronounced peak at the final frame, indicating that the model’s predictions relied predominantly on information from the frames shortly prior to the event. The average normalized attention weight was markedly lower in the earlier frames, at $0.62$ in the second-to-last frame and $0.22$ at the $10$th frame before the event. This suggests that, for mitosis, the most discriminative information was concentrated in the final frames of each trajectory.

In contrast, apoptotic sequences exhibited a broader and more gradual temporal distribution of attention. Notably, the maximum average attention was observed not at the final frame, but at the $5$th frame preceding the event. Elevated attention weights persisted across a substantial temporal range, with a gradual decline towards earlier frames. The final frame before the event retained a relatively high normalized attention weight of $0.88$, while $30$ frames before the event still exhibited an average attention weight of approximately $0.42$. Attention weights decreased more markedly only beyond this range, reaching $0.07$ at the $45$th frame prior to the event. These findings indicate that, for apoptotic sequences, informative signals were distributed across a wider temporal context rather than being confined to the immediate vicinity of the final event.

\subsubsection{Predictive Features}
\label{subsubsec:experiments:explainability:features}

Having identified frames important for prediction, we next investigated which morphological and intensity-based features distinguish high-attention frames from less informative ones. Feature quantification and statistical testing followed the procedure outlined in Sec.~\ref{subsec:methodology:explainability}. Not all cells were correctly segmented. The frames with missing segmentation were discarded from the feature analysis.

\begin{figure*}
    \centering
    \includegraphics[width=\linewidth]{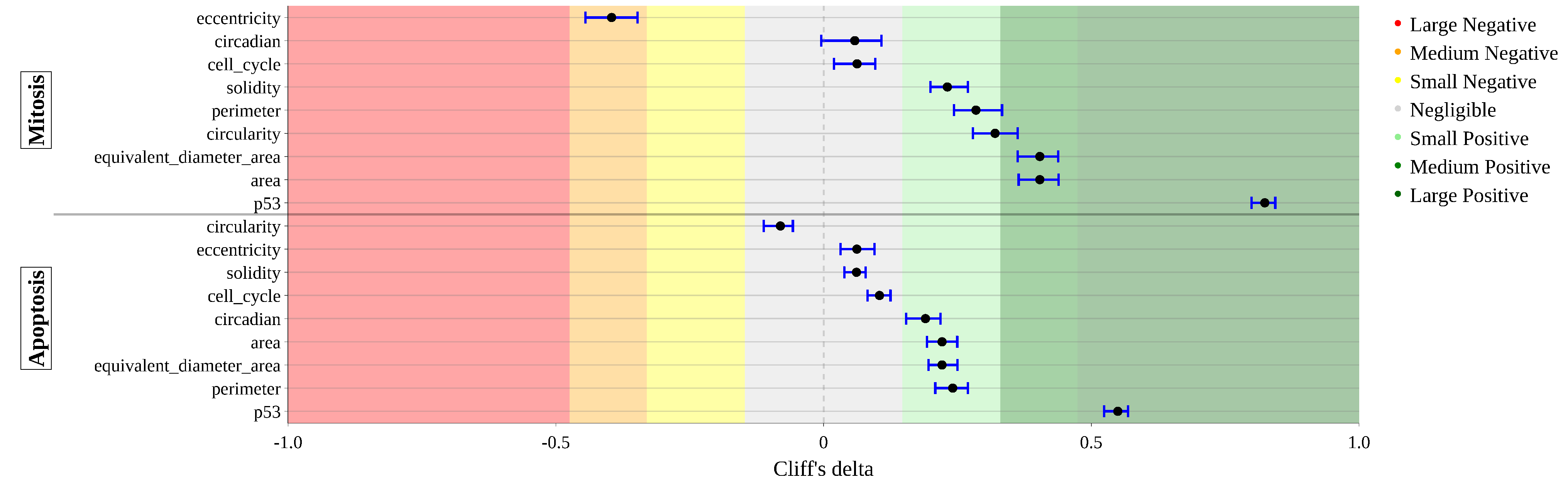}
    \caption{Effect size analysis based on Cliff’s delta for handcrafted cell features. 
    For each feature, the median Cliff’s delta and its 95\% confidence interval are shown separately for mitosis and apoptosis. 
    Positive values indicate higher feature values in high-attention frames compared to low-attention frames, whereas negative values indicate the opposite. 
    The background shading denotes commonly used effect size categories (negligible, small, medium, and large effects), facilitating interpretation of the magnitude and direction of the observed differences.}
    \label{fig:features_results}
\end{figure*}

We used permutation test and Cliff's delta to identify significantly different features between the two groups of frames. We repeated this for all sequences and calculated the median Cliff's delta and its $95\%$ confidence interval to determine the direction of the feature change. The results are shown in Fig.~\ref{fig:features_results}. 

\begin{figure*}[t]
    \centering
    \includegraphics[width=.8\linewidth]{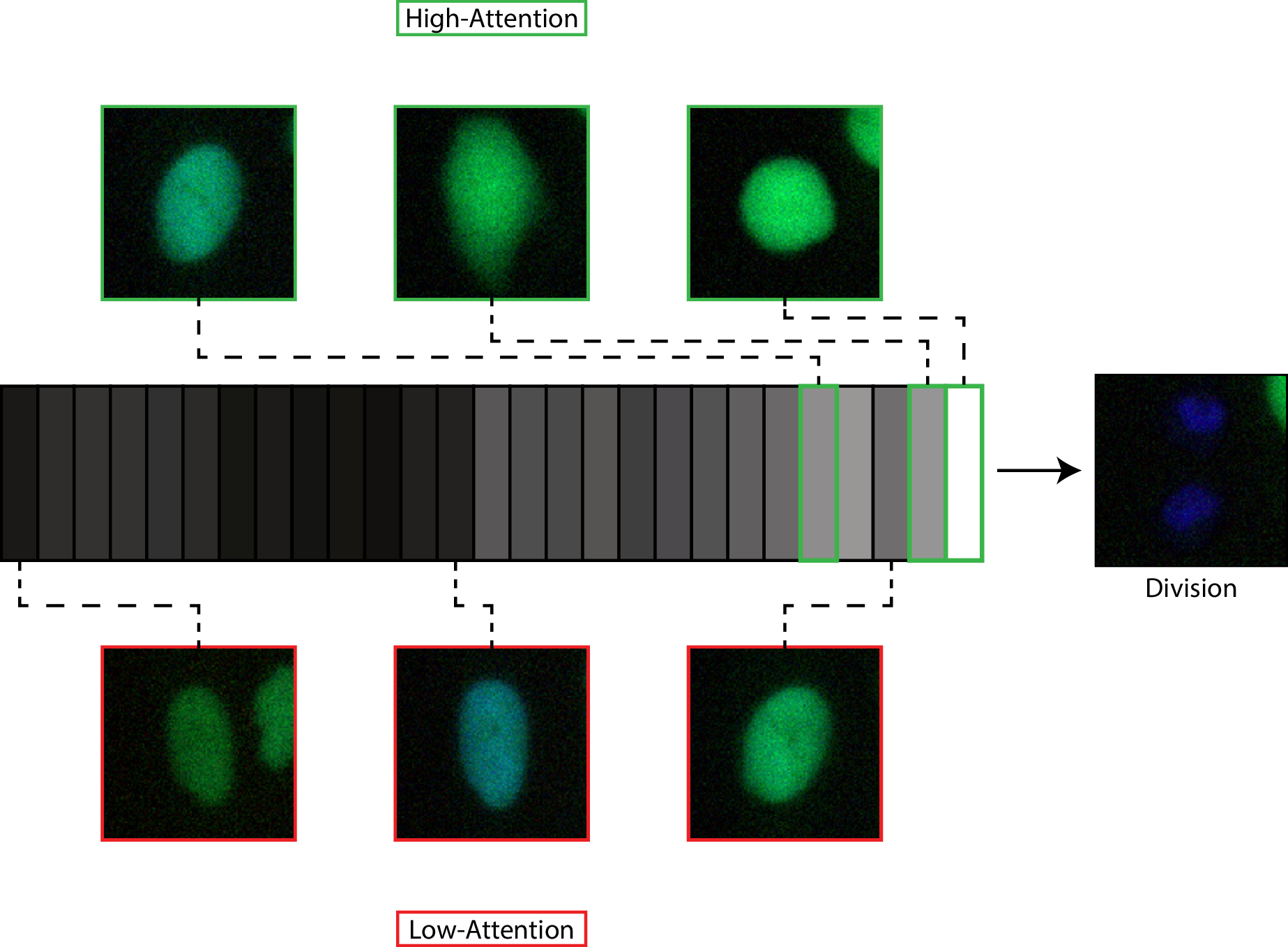}
    \caption{Illustrative mitotic trajectory with corresponding temporal attention weights. 
    The sequence depicts progressive cell growth followed by characteristic mitotic rounding in the frames immediately preceding division. 
    Frames assigned high attention by the Transformer model are highlighted in green, whereas representative low-attention frames are marked in red.}
    \label{fig:sequence_example}
\end{figure*}

For mitotic sequences, several features exhibited consistent differences between high- and low-attention frames. Most notably, the p53 channel showed a pronounced positive effect, indicating higher p53 intensity in frames that received high attention. Area-related features, including cell area and equivalent diameter, exhibited medium positive effects, while perimeter showed a smaller positive effect. Circularity displayed a small to medium positive effect, solidity a small positive effect, and eccentricity a medium negative effect. This pattern suggests that frames predictive of division were characterized by cells that appear larger and more rounded. Fig.~\ref{fig:sequence_example} shows a mitotic sequence where a progressive increase in cell size followed by pronounced mitotic rounding toward the end can be observed.

For apoptotic sequences, small positive effects were observed for equivalent diameter area, cell area, perimeter, and circadian features, indicating modest but consistent differences between high- and low-attention frames across these properties. In contrast to mitosis, no pronounced difference between high- and low-attention frames were observed for circularity and eccentricity in apoptotic sequences. This suggests that while high-attention frames in apoptotic sequences were also characterized by larger cells, they were not necessarily round, but instead retain their original shape. Notably, p53 again exhibited a large positive Cliff’s delta median, highlighting elevated p53 intensity as a dominant characteristic of frames important for apoptosis prediction. Interestingly, high-attention frames were also characterized by elevated intensity in the circadian reporter channel. This is in line with previous studies showing that the circadian reporter predicts apoptosis at the single-cell level \cite{ector2024time,ector2026circadian}. 

Finally, for both mitotic and apoptotic sequences, no significant differences were observed in the cell-cycle channel expression, which marks the progression through the specific cell-cycle phases. This suggests that cell-cycle progression was not strongly associated with the features emphasized by the model in this analysis.

\section{Discussion}
\label{sec:discussion}

The ability to predict cell fate from microscopy videos transforms time-lapse imaging from a descriptive tool into a quantitative framework for identifying early determinants of drug response, stratifying heterogeneous populations, and guiding mechanistically informed therapeutic strategies. Here we present our computational approach for cell fate prediction from raw video data of cell trajectories. Our Transformer model exploits not only static visual information but also its temporal dynamics in the prediction. Together with the model we design a thorough explainability pipeline pointing to the video frames where the predictive information is located and to the statistically significant properties of cells in these video frames.

Our first finding is that cell fate can be reliably predicted even $10$ h prior to the cell's death or division. This prediction is performed on raw video data that encodes cell morphology, presence of three dynamically expressed molecular markers and their change in time. The fact that such input allows us for cell fate prediction with high accuracy suggests that this input provides the right information for this task. Importantly, the model does not depend exclusively on the final pre-event frame but integrates signals from preceding frames throughout the sequence. 

Secondly, through a systematic explainability analysis of the model, we point to the video frames and cellular features that are exploited by the model in prediction. By combining aggregated attention distributions, temporal truncation analyses, and attention-guided feature comparisons, we observed systematic differences in how the model processed mitotic and apoptotic sequences. In the following, we discuss these observations in relation to cell fate dynamics and model explainability.

\subsection{Attention Focus}
\label{sec:discussion:attention_focus}

Differences in the temporal distribution of attention weights suggest that the model relied on distinct temporal cues when predicting mitotic versus apoptotic outcomes. In mitotic sequences, attention was almost exclusively concentrated towards the end of the sequence. This indicates that the model primarily leveraged late-stage visual cues to identify mitosis, potentially corresponding to abrupt and short-lasting changes in cell morphology or changes in expression of the imaged proteins that occurred just before division. These changes might involve rounding of the cell and fluctuation of nuclear fluorescence signals during rounding and after nuclear envelope breakdown \cite{taubenberger2020mechanics,matthews2020oncogenic,lancaster2013mitotic}. 
This is also supported by our results shown in Fig.~\ref{fig:results:truncation}d, that indicate that the model already achieved a high recall in mitosis classification based on the last $10$ frames only (recall: $0.9587$).

In contrast, apoptotic sequences exhibited a broader distribution of attention over time. Our analysis suggests that apoptotic cell death is preceded by cellular processes that emerge substantially earlier in time. Rather than concentrated near the end of the sequence, attention was distributed across a wider temporal range, with many frames showing high relevance well before cell death. We hypothesize that predictive information for apoptosis emerges gradually, such that early morphological or intensity changes already contribute to the model’s decision. This interpretation is further supported by the truncation experiment in Fig.~\ref{fig:results:truncation}d, which showed that restricting the model to only the final frames was insufficient to achieve strong performance for apoptotic sequences.

\subsection{Attention-Guided Morphological Features}
\label{sec:discussion:features}
Using our explainability analysis we point to visual features that allow the model to make a correct prediction. Our systematic, statistical analysis has pointed to mostly morphological features as well as p53 expression that are indicative of the cell outcome. Our method applied to more comprehensive cancer cell imaging data can allow to identify early determinants of drug resistance and sensitivity. Based on the identified features resistant cells can be isolated and used to guide prospective intervention — e.g., by applying secondary drugs, or perturbing signaling pathways at time points when cells are most vulnerable. Such cells can be additionally characterized using e.g., single cell transcriptomics in the particular time points when their fate is determined.

\paragraph{Mitosis}
For cells undergoing mitosis, high-attention frames contained cells with increased size, larger area, equivalent diameter, and perimeter measurements. This growth was accompanied by a decrease in eccentricity together with an increased circularity, indicating that cells progressively expanded, becoming larger and rounder. This transition from elongated to more circular shapes is consistent with known mitotic morphology, where cells undergo rounding prior to division, and aligns with previous observations in live-cell imaging studies (\cite{matthews2020oncogenic,lancaster2013mitotic}).
To predict mitosis, therefore, the Transformer primarily relies on morphological cues associated with cell rounding that occur shortly before cell division. 

Interestingly, intensity changes in p53 were also highlighted by the model as predictive of mitosis. p53 is frequently located in centrosomes during cell division and acts as a sensor for the mitotic surveillance pathway \cite{contadini2019p53}. Furthermore, p53 plays a role in the detection of mitotic extension, preventing daughter cells resulting from abnormally long mitosis from progressing further in the cell cycle \cite{meitinger2024control}. Among the myriad of functions p53 has, it is plausible that high-attention patterns in mitotic sequences relate to the role of p53 in genomic integrity. Increasing evidence also suggests that p53 signaling dynamics, such as signal duration, amplitude, or accumulation rate, encode biologically meaningful information that influences cellular outcomes \cite{purvis2013encoding}. Cells that resume proliferation after DNA damage do this via escape from p53-mediated arrest through characteristic transient changes in p53 activity shortly before division \cite{reyes2018fluctuations}. These findings underline the advantage of our approach, by processing the entire video sequence, our model captures not only snapshots of p53 activity but its dynamics across the cell cycle.

\paragraph{Apoptosis}
In contrast to mitosis, the attention patterns associated with apoptotic outcomes were more broadly distributed in time. Rather than pinpointing a consistent temporal ordering of predictive events, attention in apoptotic sequences suggests the presence of informative frames across the cell's lifespan. Frames receiving high attention were characterized by increased cell size relative to low-attention frames within the same sequence. In contrast to mitosis, however, this increase in size was not necessarily confined to the final frames preceding the event and involved a decrease in cell circularity. These morphological changes involve cell stretching rather than rounding and can occur earlier in a cell's trajectory.

In addition, high-attention frames in apoptotic sequences exhibited elevated p53 intensity, highlighting p53-related signals as an important contributor to the model’s predictions. It is well established that following DNA damage, p53 signaling is activated and plays a central role in coordinating cell fate decisions, including cell-cycle arrest, DNA repair, and the initiation of apoptosis \cite{zhang2009cell,speidel2015role,boon2024dna,gutu2023p53}. Sustained p53 activation can occur several hours before an apoptotic event \cite{zhang2011two,Chen2016p53}. We therefore speculate that prolonged p53-related intensity changes in apoptotic sequences may constitute one of the early cues captured by the model and reflected in the broader temporal distribution of attention. Notably, and in contrast to mitotic sequences, high-attention frames for apoptosis also showed increased intensity in the circadian channel. This suggests that circadian-associated dynamics, together with sustained p53-related signals and moderate morphological changes, may provide early and distributed cues that the model leverages to identify cells destined for apoptosis. Overall, the distinct temporal attention profiles for mitotic versus apoptotic sequences mirror the distinct p53 dynamical states that mechanistic studies have identified as fate-determining \cite{reyes2018fluctuations,Chen2016p53}. This underlines that temporal deep learning models are appropriate tools that naturally capture the phase- and amplitude-dependent information encoded in signaling dynamics - information that is invisible to static or short-window analyses.
\newline\newline
In summary, by using truncation experiments and explainability analysis we quantify how predictive information accumulates over time with longer observation window. The ability to predict cell death early in a cell’s lifetime reflects the extent to which this fate is pre-determined versus progressively acquired. Our approach enables identification of critical temporal windows during which fate becomes predictable.
\section{Conclusion}
\label{sec:conclusion}

The combination of Transformer models with attention-based interpretability provides a powerful framework for the analysis of complex video data. By operating on full temporal sequences and leveraging the attention mechanism, the proposed framework is particularly well suited to capturing dynamic cellular processes and temporally evolving molecular signals present in time-lapse microscopy data. Our results demonstrate that this approach not only enables prediction of biological phenomena but also allows the extraction and interpretation of the signals underlying these predictions. Particularly in biological systems that remain incompletely understood, the methodology presented here offers a valuable strategy for identifying informative patterns and gaining insight into the mechanistic determinants that drive specific phenotypic outcomes.
\section{Declaration of generative AI and AI-assisted technologies in the writing process}
During the preparation of this manuscript, the author used ChatGPT in order to rephrase certain sentences for improved readability. After using this tool/service, the author reviewed and edited the content as needed and takes full responsibility for the content of the publication.

\section{Acknowledgements}
Florian Bürger and Katarzyna Bozek were supported by the North Rhine-Westphalia return program (311-8.03.03.02-147635), Martim Dias Gomes was supported by the BMBF program Junior Group Consortia in Systems Medicine (01ZX1917B). Noémie Moreau was supported by the Deutsche Forschungsgemeinschaft (DFG, German Research Foundation) - Project No. 386793560. Florian Bürger, Katarzyna Bozek, Martim Dias Gomes, and Noémie Moreau were hosted by the Center for Molecular Medicine Cologne. We thank the IT Center of the University of Cologne (ITCC) for providing support and computing time.

\bibliographystyle{elsarticle-harv} 
\bibliography{literature.bib}

\end{document}